\documentclass{article}
\usepackage{spconf,amsmath,graphicx,epstopdf}

\def \b {\mathbf}
\def \tb {\textbf}

\DeclareMathOperator*{\argmax}{arg\,max}

\title{Speaker Cluster-Based Speaker Adaptive Training for Deep Neural Network Acoustic Modeling}
\name{Wei Chu \quad \quad \quad Ruxin Chen\thanks{We would like to thank the authors of Kaldi speech recognition toolkit for making it opensource.}}
\address{Sony Computer Entertainment America\\
		2207 Bridgepointe Pkwy, San Mateo, CA 94404\\
		\texttt{\{wei\_chu, ruxin\_chen\}@playstation.sony.com}}

\begin{document}
%
\maketitle
\begin{abstract}
A speaker cluster-based speaker adaptive training (SAT) method under deep neural network-hidden Markov model (DNN-HMM) framework is presented in this paper. During training, speakers that are acoustically adjacent to each other are hierarchically clustered using an i-vector based distance metric. DNNs with speaker dependent layers are then adaptively trained for each cluster of speakers. Before decoding starts, an unseen speaker in test set is matched to the closest speaker cluster through comparing i-vector based distances. The previously trained DNN of the matched speaker cluster is used for decoding utterances of the test speaker.  The performance of the proposed method on a large vocabulary spontaneous speech recognition task is evaluated on a training set of with 1500 hours of speech, and a test set of 24 speakers with 1774 utterances. Comparing to a speaker independent DNN with a baseline word error rate of 11.6\%, a relative  6.8\% reduction in word error rate is observed from the proposed method.
 
\end{abstract}
\section{Introduction}
\label{sec:intro}

Speaker variation is one of the important factors that affect the performance of a practical large vocabulary spontaneous speech recognizer, researchers have spent a significant amount of efforts in exploring effective adaptation methods. 

During the era in which the Gaussian mixture model-hidden Markov model (GMM-HMM) framework is predominant for acoustic modeling, efforts can be mainly categorized into two classes: unsupervised adaptation during decoding, and speaker adaptive training (SAT)~\cite{Anastasakos96acompact}. In unsupervised adaptation, e.g. fMLLR~\cite{Gales98maximumlikelihood}, speaker-specific feature transformation is obtained through accumulating statistics from first-pass decoding, then a second-pass decoding is conducted on the transformed feature. SAT tries to obtain a compact model by decoupling phonetic and speaker variation during training, and to estimate a speaker-specific feature transformation during decoding. It is possible for SAT to finish the decoding in one pass, which improves the speed of decoding compared to unsupervised adaptation. There are two styles of speaker-specific feature transformations: global style and regression class style. In global style, all Gaussians share the same transformation. While in regression class style, a regression class tree which can statistically determine the number of classes is used for clustering Gaussians, and only Gaussians that are clustered together share the same transformation. Furthermore, the transformations of different clusters can be interpolated in a cluster adaptive training framework~\cite{gales2000cluster}.

When the state-of-the-art acoustic modeling technology evolves to DNN, researchers keep trying unsupervised adaptation and SAT~\cite{yao2012adaptation}. Other than millions of Gaussians to be adapted in the GMM-HMM framework, there is only one DNN to be adapted. Because of the change in model topology, the regression class-style adaption is no longer valid. While the difficulty of the adaptation for DNNs still remains. Because that there may be too many parameters to be updated, regarding that available speaker data for adaptation are usually not comparatively enough.

In the direction of unsupervised adaptation for DNN, researchers tended to adapt only the affine transform plus bias or only bias in one layer~\cite{yao2012adaptation}~\cite{kumar2015intermediate}, which still have millions or a few thousands of parameters to be updated. To reduce the number of parameters involved in adaptation, an SVD layer can be inserted ~\cite{xue2013restructuring}, or a linear transformation or a bias to the input feature can be estimated~\cite{li2010comparison}, or one can only adapt activation functions of neurons in just one of the all layers, which only have thousands of parameters~\cite{pawel2014learning}. 


In the direction of SAT for DNN, researchers are taking advantage of the availability of a large amount of transcribed data. i-vector based methods are commonly used. An speaker-specific i-vector feature vector can be appended to each frame of existing feature vector as an auxiliary feature vector~\cite{gupta2014vector}~\cite{garimella2015ivector}. Proper combination weights between original feature vector and i-vector feature vector are supposed to be obtained through back propagation (BP) algorithm during DNN training. The i-vector-based approach can increase the computation load during decoding, not only because of the time spent on i-vector extraction, but also because of the increase in the size of both feature and model. It has been discovered that reducing the durations of test utterances can hurt the performance of a i-vector-based speaker verification system~\cite{sarkar2012study}. In real-word scenario, when there is no sufficient amount of available speech from a test speaker, it may be difficult for the calculated i-vector features to robustly represent the characteristics of the speaker.  There are also methods which do not necessarily require i-vector features~\cite{xue2014direct}~\cite{tan2015cluster}. For example, cluster adaptive training can be used to obtain multiple affine transforms in a DNN layer, then to estimate speaker-specific interpolation coefficients to fuse the affine transforms for an unseen testing speaker~\cite{tan2015cluster}.

Inspired by the  regression class tree in fMLLR and cluster adaptive training methods, the proposed method clusters speakers from training set that are close in i-vector feature space together. The phonetic and speaker variation of speakers in each cluster are modeled by a shared speaker independent (SI) DNN plus a speaker cluster-specific layer. On test set, it is assumed that an unseen speaker can be regarded as similar to certain speakers which have appeared in the training set, i.e. close to certain speaker cluster in i-vector feature space. The DNN of the speaker cluster whose i-vector is the closest to the test speaker is used for decoding. No second pass decoding is needed.

The details of the algorithm and the experimental results of the proposed speaker cluster-based SAT-DNN training are shown in the following sections.

\section{Speaker Cluster-based SAT-DNN}

The proposed speaker cluster-based SAT-DNN training and decoding procedure is described in the following subsections.

\subsection{SI-DNN Training}
Using standard training recipe for a speaker independent (SI) feed forward DNN. A DNN with a set of parameters, i.e. $\b \Lambda$, is regarded as a function for transforming an input $\b x$ to an output $\b y$;
\begin{equation}
\b y = f(\b x; \b \Lambda)
\label{eq1}
\end{equation}
Suppose the DNN is composed of $L$ layers from bottom to top: $\b \Lambda = \{\b \Lambda_1, \cdots,  \b \Lambda_L \}$. 
In the following, when it is mentioned that $\b \Lambda_l$ is updated,  it means that the affine transform and bias of $l_{\text{th}}$ layer are updated through BP algorithm. 

\subsection{Speaker Clustering}

Two persons may sound acoustically similar to each other in certain speaking style which include accent, talking speed, and so on. If a training set can be infinitely large, one unseen speaker in a test set can always be matched to another speaker in the training set who has a similar speaking style. The decoding of the utterances from this unseen speaker can benefit from using the speaker dependent (SD) model of the matched speaker which has already been trained. Whereas, due to the limitation on the size of training sets, and the limitation on the allowable time to search for matching speakers during decoding, it is not quite practical to train a set of SD models for each speakers appeared in the training set. It is natural to consider clustering the speakers of similar speaking style together.

To cluster speakers, the i-vector has been used and been proved effective~\cite{khoury2014hierarchical}. In this paper, define the distance between speaker $S_i$ and $S_j$ as:

\begin{equation}
D(S_i || S_j) = \langle \b {i}^{S_i}, \b {i}^{S_j} \rangle
\label{eq1}
\end{equation}
where $\langle \cdot, \cdot \rangle$ denotes inner product, $\b {i}^{S_i}$ and $\b {i}^{S_j}$ denote the normalized i-vectors obtained from all available data of speaker $S_i$ and $S_j$, respectively. After defining the distance, Ward's method can be used as the criterion to cluster the speakers into several clusters on the training set~\cite{khoury2014hierarchical}. i-vectors are then calculated for each cluster $C_k$, i.e. $ \b {i}^{C_k}$, using all the data from speakers in the cluster.  

Usually, a training set can have hundreds or thousands of speakers. This paper empirically tested the effect of the number of clusters on the performance of the DNN-HMM system. Analysis results will be shown in the experimental section.

\subsection{SAT-DNN Training}

Select one layer from the SI-DNN as the SD layer and mark layers other than the SD layer as other SI layers. Initially, duplicate $N$ copies of the previously trained SI-DNN for $N$ speaker clusters. Let $N$ other SI layers in the DNN, i.e. $\b \Lambda^C_{\texttt{SI\_OTHER}}$, $C=1,\cdots, N$, share the same set of parameters, i.e. $\b \Lambda_{\texttt{SI\_OTHER}}$. The $N$ SD layers in the DNN, i.e. $\b \Lambda^C_{\texttt{SD}}$, $C=1,\cdots, N$, does not share parameters. Now the output can be calculated as:
\begin{equation}
\b y =f(\b x; \{\b \Lambda_\texttt{SI\_OTHER}, \b \Lambda_\texttt{SD}^C\}), \quad \text{if} \enspace \b x \in C
\end{equation}

For each speaker cluster, run BP algorithm to update only the SD layer while keeping other SI layers unchanged, which is performed on only the training examples from the speakers in the cluster. Then keep the SD layer of each speaker cluster unchanged, update the shared parameters of other SI layers using the training examples from all speakers. This training process can be repeated for iterations, until it converges, i.e. the increase in the value of cross entropy-based objective function falls below an empirically set threshold. 

\subsection{SAT-DNN Decoding}

After finishing the training procedure described in the previous subsection, during decoding time, an unseen speaker $S$ is matched to the speaker cluster $C^*$ as follows:
\begin{equation}
C^* = \argmax_{C} <\b {i}^{S}, \b {i}^{C}>
\label{eq2}
\end{equation}
where $\b i^S$ denotes the i-vector of the speaker $S$, which can be calculated on all or only one of her/his utterances,  $ \b {i}^{C}$ denotes the i-vector of speaker cluster $C$. The SD-DNN $\{\b \Lambda_\texttt{SI\_OTHER}, \b \Lambda_\texttt{SD}^{C^*}\}$ is selected for decoding.

This paper uses a similar iterative training scheme for the SD and SI layers in DNN as Ochiai et al.~\cite{ochiai2014speaker}. This paper trains SAT models on training, while Ochiai et al. trained SAT models on part of the test set in a cross validation fashion. Instead of choosing one transform from a set of speaker-specific affine transforms to form a SD-DNN as proposed in this paper, Ochiai et al. unsupervisedly trained a SD layer contained only one affine transform for the test speaker. This paper performs only one pass decoding, while Ochiai et al. ran extra rounds of adaptation to the SD layer before running the final pass of decoding. 

Tan et al.~\cite{tan2015cluster} retained cluster adaptive training (CAT) layer and SI layers in final trained DNN. The CAT layer is composed of a set of canonical affine transforms plus bias which are learned through BP algorithm, which is similar to multiple speaker-specific affine transforms plus bias in SD layer in this paper. Tan et al. ran first pass decoding using an SI-DNN to obtain state-level alignments, then estimated speaker-specific interpolation coefficients to combine the canonical transforms for the CAT layer before starting second pass decoding. This paper selects only one among all speaker-cluster-specific transforms to form the final SD-DNN. The method proposed in this paper does not require two pass decoding. Interestingly, in Section 2.2 of Tan et al.'s work, the concept of speaker clustering was mentioned as a possible way for initializing canonical transforms in CAT layer, although the direction was not explored further.



\section{Experiments}

The proposed speaker cluster-based SAT-DNN-HMM system is evaluated on an English spontaneous speech recognition task. The training set has two portions. The first portion is a general training set, which consists of 1400 hrs of WSJ-SI284, Switchboard, half of the Fisher (no full Fisher due to legal reasons) corpora. The second portion is a in-domain training set, which consists of 35 hrs of self-collected in-domain data from 277 speakers. The development set contains 22 speakers with 1288 in-domain utterances, which is only used for tuning language model weights during decoding. The test set contains 24 speakers with 1774 in-domain utterances. The averaged number of word in each utterance is 15.

In this paper, C1 to C12 of MFCCs plus logarithm energy feature are used as static feature. The static feature with its first and seconder order derivatives are stacked together to form a 39 dimensional feature. Adjacent 9 frames of static plus dynamic features are stacked together to form 351 dimensional long-span features. Then LDA+MLLT is used to reduced its dimension to 40. A GMM-HMM system which uses the LDA+MLLTed features and SAT and fMLLR adaptation methods is firstly trained to obtain the alignment from training data for the subsequent DNN training. The number of senones is 8891. 3-gram language model is trained on the text of the general training set, then adapted according to the text of the in-domain training set.

The opensource toolkit for speech recognition and processing, Kaldi~\cite{Povey_ASRU2011}, is used for both GMM and HMM training. Maxout network with $p = 2$ is used, the input and output dimensions of pnorm component are set to 2000 and 400, respectively. The features used in DNN are obtained by stacking adjacent 9 frames LDA+MLLTed features together, i.e. a 360 dimensional feature. Cross entropy-based objective function is used.

The general SI-DNN with 7 layers is trained purely on the general training set first. During training, the initial and final learning rates are set to 0.02 and 0.002, respectively. The minibatch size is set to 512. Data parallelization is used to accelerate the training speed. Note that when the results of decoding using the SD-DNN are presented in this paper, no afterward unsupervised adaption is performed because no significant further reduction in WER is observed.

For speaker clustering, the dimension of i-vector is set to 100. For GMM mean supervector calculation, a full covariance GMM with 512 Gaussians is trained on the in-domain training set. The extracted i-vector is normalized. Detailed speaker clustering analysis results will be shown in the following paragraphs.

For in-domain SAT-DNN, the code of Kaldi toolkit has been modified to perform the customized training proposed in this paper. The number of iteration is set to 10.  The learning rate is set to a constant value, 0.1. The locations of SD layers is set to the first layer. The lowest WER of SAT-DNN is achieved under this configuration. A comparison of the WERs when using the general SI-DNN, the in-domain SI-DNN, and the in-domain SAT-DNN for decoding on the test set is shown in Table 1. The proposed SAT-DNN yields a WER of 10.83\%, which results in a 6.8\% relative reduction compared to the WER of in-domain SI-DNN. It is also observed that after the first iteration, the fluctuation in WERs across iterations is less than 0.05\%.

\begin{table}
\begin{center}
\caption{WERs (\%) of different DNNs on test set. The general SI-DNN is trained on the general training set. The in-domain SI-DNN is trained on additional in-domain data using the general SI-DNN as initial model. The proposed SAT-DNN in this paper is adaptively trained on in-domain data using the in-domain SI-DNN as initial model. }

\begin{tabular}{|c||c|c|c|}
  \hline
   General SI-DNN & 13.67 \\ \hline
In-domain SI-DNN & 11.62 \\ \hline
In-domain SAT-DNN & 10.83 \\ \hline
\end{tabular}
\vspace*{-5mm}
\end{center}
\label{tab:si_sat}
\end{table}

\begin{table}
\begin{center}
\caption{WERs (\%) of when changing number of speakers in SAT-DNN on the test set.`Cluster=i' denotes that the number of speaker cluster in the SAT-DNN is set to i. When changing the  location of SD layer in the SAT-DNN, the WER also changes accordingly.}
\begin{tabular}{|c||c|c|c|}

\hline
 & \multicolumn{3}{|c|}{SD Layer's Location} \\ \cline{2-4}

  & 1st & 4th & 7th \\ \hline\hline
SAT-DNN, Cluster=5  & 11.00 & 10.95 & 10.97  \\ \cline{2-4}
\quad \quad \quad \quad \quad Cluster=10   & \tb{10.83} & 10.85 & 10.92  \\ \cline{2-4}
\quad \quad \quad \quad \quad Cluster=20  & 11.01 & 10.99 & 11.12  \\ \hline
\end{tabular}
\vspace*{-5mm}
\end{center}
\label{tab:comp}
\end{table}

As shown in Table 2, the number of speaker clusters (5, 10, or 20) and the location of SD layer (1st, 4th, or 7th) can both affect the final WER. When the number of clusters is set to 10, and the SD layer is set to the first layer, the best performance is achieved. When fixing the number of clusters, different locations of SD layer do not affect WER much (less than 0.1 \%). It can also been seen that the WERs fluctuate in a small range when these parameters change.

When the number of speaker clusters increases, it may cause the amount of speaker data for adaption to decrease, resulting in an overfit SD layer. When the number of speaker clusters decreases, the speaker variation from different speakers may not be well modeled because there may be too many speakers clustered into a same cluster. 

The speaker clustering results on the in-domain training set are shown in Table 3. It can be seen that when the number of clusters changes, there is no cluster with only one or two speakers appeared, which may avoid the overfitting problem. Take the 10 cluster case for example, the minimum number of speakers in a cluster is 14, which contains 722 utterances with a total duration of 1.4 hours of speech. Compared to unsupervised adaptation scenarios, the speaker cluster-based SAT can have more data to adapt an SD layer, even if the number of speakers in that cluster is small. Certain clusters can end up with a large amount of speakers after clustering, though. For example, under 10 cluster case, the maximum number of speakers in a cluster reached 57 speakers, which means the cluster is associated with 7 hours speaker adaptation data.

\begin{table}
\begin{center}
\caption{Speaker clustering results on the in-domain training set with 277 speakers in total. `Cluster=i' denotes that the number of speaker cluster in the SAT-DNN is set to i. min/max/avg denotes the minimum/maximum/averaged number of speakers in a cluster, respectively.}
\begin{tabular}{|c||c|c|c|}
\hline
  & \multicolumn{3}{|c|}{The Number of Speakers in Clusters} \\ \hline
  & min &  max  & avg  \\ \hline\hline
Cluster = 5  & 23 & 126 & 55.4 \\ \hline
Cluster = 10  & 14 & 57 & 27.7 \\ \hline
Cluster = 20  & 7 & 23 & 13.9 \\ \hline
\end{tabular}
\vspace*{-5mm}
\end{center}
\label{tab:sc}
\end{table}

The performance of the speaker clusters is also analyzed. A 5-fold cross validation set is created on the in-domain training set. For a speaker in a validation set, if it is matched to a speaker cluster in the corresponding training set according to Eq.~\ref{eq2}, and the speaker cluster contains this speaker, it is considered as a successful speaker cluster matching. An accuracy metric, Speaker Cluster Matching Accuracy (SCMA), on a validation set is defined as 
\begin{equation}
\text{SCMA} = \frac{\text{\# of successfully matched speakers}}{\text{\# of total speakers}} \times 100\%
\end{equation}
Note that when a speaker is matched, it does not have to know which speaker in that matched cluster it is matched to. Speaker cluster matching is different from speaker recognition.
  
When there are 10 speaker clusters in training set, the averaged SCMAs on all validation sets is 83.5\%. SCMA cannot be calculated on the test set, for the speakers are unseen speakers. Although the SCMA analysis is not available for speakers from a test set, one can still imagine that when a training set is infinitely large in size, i.e., has a full coverage of all available speakers, the speakers on a test set can be viewed as seen speakers. The higher SCMA is, the more likely the test speaker can be matched to the correct speaker cluster. Or, it can be said that a test speaker is more likely to use a correctly matched SD model in decoding, which will likely result in a reduction in WER. Therefore, improving speaker cluster matching accuracy might help to reduce WER when the size of the training set is large.

In this paper, unsupervised adaptation to the SD layer is also performed after first pass decoding. No significant further reduction in WER is observed in the following second pass decoding when the model used in first pass decoding is the in-domain SAT-DNN. One possible explanation is that the data from 70 untranscribed utterances of each test speaker are not enough to reliably further adapt the 8$\times 10^5$ parameters in the affine transform plus bias, or 2000 parameters in the bias, in the SD layer. The reason why the SAT-DNN works might be because that the proposed model can take advantage of large amount of transcribed data to train speaker cluster-specific models. And also the speaker cluster matching is effective in selecting the proper matched SD model, when an adequate amount of clusters is chosen, and a sufficient amount of test speaker data is available for extracting reliable i-vector features.

\section{Conclusions}

In this paper, a speaker cluster-based SAT-DNN framework is proposed and proven. i-vector based distance metric is used to cluster the speakers in the training set together. An SAT-DNN with multiple canonical transforms in SD layer is trained on the speaker clusters. An unseen speaker in test set is matched to the closest speaker cluster through comparing i-vector based distances. The proposed method finishes the decoding in one pass using the SD-DNN associated with the matched speaker cluster. A relative  6.8\% reduction in WER is observed compared to an SI-DNN model.

%

%
%
%
%
%

\vfill\pagebreak


\bibliographystyle{IEEEbib}
\bibliography{strings,refs}

\begin{thebibliography}{10}

\bibitem{Anastasakos96acompact}
Tasos Anastasakos, John Mcdonough, Richard Schwartz, and John Makhoul,
\newblock ``A compact model for speaker-adaptive training,''
\newblock in {\em in Proc. ICSLP}, 1996, pp. 1137--1140.

\bibitem{Gales98maximumlikelihood}
M.J.F. Gales,
\newblock ``Maximum likelihood linear transformations for hmm-based speech
  recognition,''
\newblock {\em Computer Speech and Language}, vol. 12, pp. 75--98, 1998.

\bibitem{gales2000cluster}
Mark~JF Gales,
\newblock ``Cluster adaptive training of hidden markov models,''
\newblock {\em Speech and Audio Processing, IEEE Transactions on}, vol. 8, no.
  4, pp. 417--428, 2000.

\bibitem{yao2012adaptation}
Kaisheng Yao, Dong Yu, Frank Seide, Hang Su, Li~Deng, and Yifan Gong,
\newblock ``Adaptation of context-dependent deep neural networks for automatic
  speech recognition,''
\newblock in {\em Spoken Language Technology Workshop (SLT), 2012 IEEE}. IEEE,
  2012, pp. 366--369.

\bibitem{kumar2015intermediate}
Kshitiz Kumar, Chaojun Liu, Kaisheng Yao, and Yifan Gong,
\newblock ``Intermediate-layer dnn adaptation for offline and session-based
  iterative speaker adaptation,''
\newblock in {\em Interspeech}, 2015.

\bibitem{xue2013restructuring}
Jian Xue, Jinyu Li, and Yifan Gong,
\newblock ``Restructuring of deep neural network acoustic models with singular
  value decomposition.,''
\newblock in {\em INTERSPEECH}, 2013, pp. 2365--2369.

\bibitem{li2010comparison}
Bo~Li and Khe~Chai Sim,
\newblock ``Comparison of discriminative input and output transformations for
  speaker adaptation in the hybrid nn/hmm systems,''
\newblock in {\em Interspeech}, 2010, pp. 526--529.

\bibitem{pawel2014learning}
Steve Renals and Pawel Swietojanski,
\newblock ``Learning hidden unit contributions for unsupervised speaker
  adaptation of neural network acoustic models,''
\newblock in {\em IEEE Spoken Language Technology}, 2014.

\bibitem{gupta2014vector}
Vishwa Gupta, Patrick Kenny, Pierre Ouellet, and Themos Stafylakis,
\newblock ``I-vector-based speaker adaptation of deep neural networks for
  french broadcast audio transcription,''
\newblock in {\em Acoustics, Speech and Signal Processing (ICASSP), 2014 IEEE
  International Conference on}, 2014, pp. 6334--6338.

\bibitem{garimella2015ivector}
Sri Garimella, Arindam Mandal, Nikko Strom, Bjorn Hoffmeister, Spyros
  Matsoukas, and Sree Hari~Krishnan Parthasarathi,
\newblock ``Robust i-vector based adaptation of dnn acoustic model for speech
  recognition,''
\newblock in {\em Interspeech}, 2015, pp. 2877--2881.

\bibitem{sarkar2012study}
Achintya~Kumar Sarkar, Driss Matrouf, Pierre-Michel Bousquet, and
  Jean-Fran{\c{c}}ois Bonastre,
\newblock ``Study of the effect of i-vector modeling on short and mismatch
  utterance duration for speaker verification.,''
\newblock in {\em INTERSPEECH}, 2012.

\bibitem{xue2014direct}
Shaofei Xue, Ossama Abdel-Hamid, Hui Jiang, and Lirong Dai,
\newblock ``Direct adaptation of hybrid dnn/hmm model for fast speaker
  adaptation in lvcsr based on speaker code,''
\newblock in {\em Acoustics, Speech and Signal Processing (ICASSP), 2014 IEEE
  International Conference on}. IEEE, 2014, pp. 6339--6343.

\bibitem{tan2015cluster}
Tian Tan, Yanmin Qian, Maofan Yin, Yimeng Zhuang, and Kai Yu,
\newblock ``Cluster adaptive training for deep neural network,''
\newblock in {\em Acoustics, Speech and Signal Processing (ICASSP), 2015 IEEE
  International Conference on}. IEEE, 2015, pp. 4325--4329.

\bibitem{khoury2014hierarchical}
Elie Khoury, Laurent El~Shafey, Marc Ferras, and S{\'e}bastien Marcel,
\newblock ``Hierarchical speaker clustering methods for the nist i-vector
  challenge,''
\newblock in {\em Odyssey: The Speaker and Language Recognition Workshop},
  2014, number EPFL-CONF-198439.

\bibitem{ochiai2014speaker}
Toshihiko Ochiai, Shodai Matsuda, Xugang Lu, Chiori Hori, and Souichi Katagiri,
\newblock ``Speaker adaptive training using deep neural networks,''
\newblock in {\em Acoustics, Speech and Signal Processing (ICASSP), 2014 IEEE
  International Conference on}. IEEE, 2014, pp. 6349--6353.

\bibitem{Povey_ASRU2011}
Daniel Povey, Arnab Ghoshal, Gilles Boulianne, Lukas Burget, Ondrej Glembek,
  Nagendra Goel, Mirko Hannemann, Petr Motlicek, Yanmin Qian, Petr Schwarz, Jan
  Silovsky, Georg Stemmer, and Karel Vesely,
\newblock ``The kaldi speech recognition toolkit,''
\newblock in {\em IEEE 2011 Workshop on Automatic Speech Recognition and
  Understanding}. Dec. 2011, IEEE Signal Processing Society,
\newblock IEEE Catalog No.: CFP11SRW-USB.

\end{thebibliography}

\end{document}